\documentclass[letterpaper, 10 pt, conference]{ieeeconf}  % Comment this line out if you need a4paper

\IEEEoverridecommandlockouts                              % This command is only needed if 
\usepackage{graphics} % for pdf, bitmapped graphics files
\usepackage{epsfig} % for postscript graphics files
\usepackage{mathptmx} % assumes new font selection scheme installed
\usepackage{times} % assumes new font selection scheme installed
\usepackage{amsmath} % assumes amsmath package installed
\usepackage{amssymb}  % assumes amsmath package installed

\usepackage{algorithm}
\usepackage{algorithmic}
\usepackage{booktabs}
\usepackage{subfig}
\usepackage{multirow}

\title{\LARGE \bf
Real-Time Evaluation of Autonomous Systems under Adversarial Attacks
}

\author{Adithya Mohan$^{1}$, Xujun Xie$^{1}$, Venkatesh Thirugnana Sambandham$^{1}$ and Torsten Schön$^{1}$% <-this % stops a space
\thanks{This work was supported by Hightech Agenda Bayern}% <-this % stops a space
\thanks{$^{1}$The authors are with AI Motion Institute,
        Technische Hochschule Ingolstadt, Germany
        {\tt\small First Name.Last Name@thi.de}}%
}

\begin{document}

\maketitle
\thispagestyle{empty}
\pagestyle{empty}

\begin{abstract}
Most evaluations of autonomous driving policies under adversarial conditions are conducted in simulation, due to cost efficiency and the absence of physical risk. However, purely virtual testing fails to capture structural inconsistencies, supervision constraints, and state-representation effects that arise in real-world data and fundamentally shape policy robustness. This work presents an offline trajectory-learning and adversarial robustness evaluation framework grounded in real-world intersection driving data. Within a controlled data contract, we train and compare three trajectory-learning paradigms: Multi-Layer Perceptron (MLP)-based Behavior Cloning (BC), Transformer-based object-tokenized BC, and inverse reinforcement learning (IRL) formulated within a Generative Adversarial Imitation Learning (GAIL) framework. Models are evaluated using Average Displacement Error (ADE) and Final Displacement Error (FDE).

Inference-time robustness is assessed by subjecting trained policies to gradient-based adversarial perturbations across multiple intersection scenarios, yielding a structured robustness evaluation matrix. Results show that state-structure design and architectural inductive biases critically influence adversarial stability, leading to markedly different robustness profiles despite comparable nominal prediction accuracy (ADE $<$ 0.08). Inference-time Projected Gradient Descent (PGD) attacks induce final displacement errors of up to approximately 8 meters. The proposed framework establishes a scalable benchmark for studying offline trajectory learning and adversarial robustness in real-world autonomous driving settings.
\end{abstract}

\begin{figure*}[h]
    \centering
    \includegraphics[width=\linewidth]{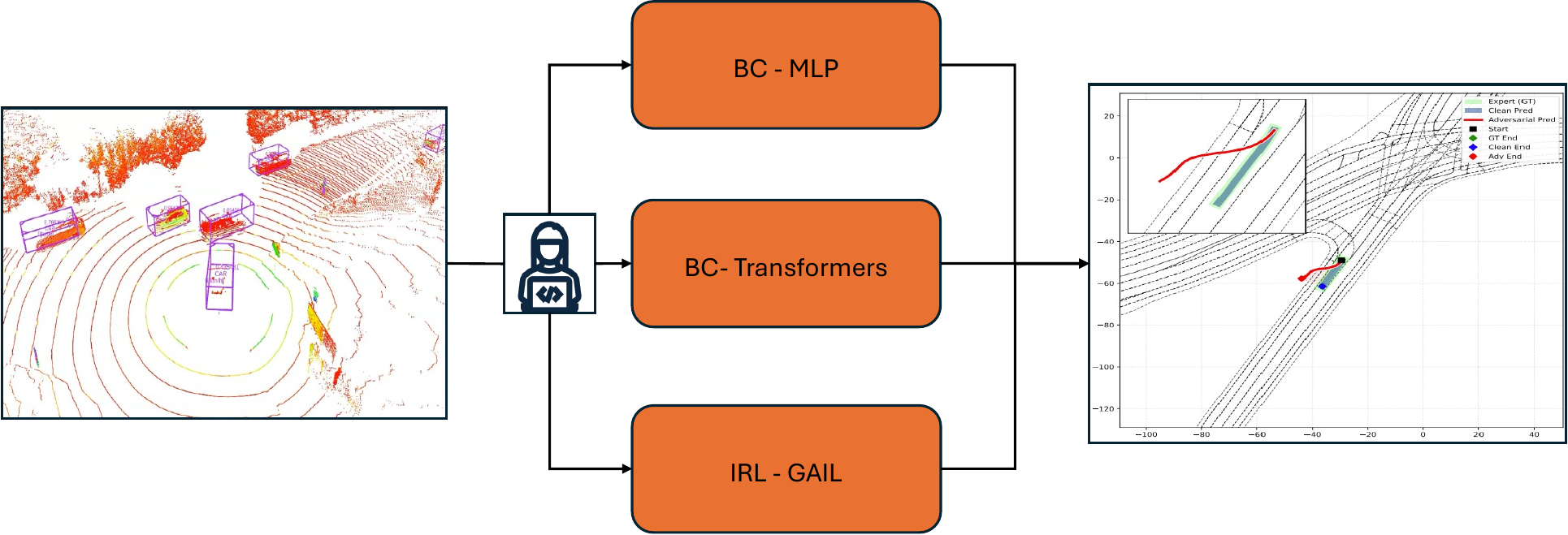}
    \caption{Real-world open-loop inference-time robustness evaluation pipeline.}
    \label{fig:Story}
\end{figure*}

%%%%%%%%%%%%%%%%%%%%%%%%%%%%%%%%%%%%%%%%%%%%%%%%%%%%%%%%%%%%%%%%%%%%%%%%%%%%%%%%
\section{Introduction}

Robustness evaluation has become a central concern in autonomous driving, particularly as learning-based policies are increasingly deployed in safety-critical settings. While adversarial testing is now standard practice in simulation, translating such evaluations to real-world driving data remains challenging \cite{dosovitskiy2017carla}. Simulation environments provide controllability and safety, but they abstract away structural characteristics of real driving logs, including annotation inconsistencies, temporal alignment constraints, and state-construction artifacts that directly affect offline trajectory learning. Fig.~\ref{fig:Story} illustrates the overall evaluation pipeline, from real-world sensing and structured state construction to open-loop inference-time robustness analysis under adversarial perturbations.

Imitation learning, especially Behavior Cloning (BC), is widely used for trajectory prediction and policy approximation in autonomous driving. More expressive formulations such as Inverse Reinforcement Learning (IRL) and Generative Adversarial Imitation Learning (GAIL) attempt to recover latent reward structures underlying expert demonstrations \cite{ho2016generative, gupta2018social}. Despite their strong nominal predictive performance, recent studies have shown that learning-based policies can exhibit brittle behavior under adversarial perturbations \cite{goodfellow2015explaining,madry2019towards}. In the context of reinforcement learning and autonomous systems, robustness vulnerabilities have been systematically analyzed in recent work \cite{mohan2026toward,mohan2025advancing,karpenahalli2025evolution}, yet controlled evaluations on structured real-world trajectory datasets remain limited.

Large-scale driving datasets such as Nuscenes \cite{caesar2020nuscenesmultimodaldatasetautonomous}, Waymo \cite{waymo}, UrbanIng-V2X \cite{sekaran2025urbaning} and DrivIng \cite{rossle2026driving} have enabled research in perception and multimodal fusion across complex intersections and traffic scenarios. However, comparatively little attention has been devoted to reproducible offline trajectory-learning pipelines that explicitly analyze inference-time adversarial robustness under unified state and supervision contracts. All experiments are conducted in an open-loop evaluation mode, where learned policies perform real-time inference on live vehicle data without influencing vehicle control. While adversarial attacks were initially studied in digital settings, recent research has demonstrated their efficacy in the physical world, where environmental conditions are dynamic. Notably, Kong et al.~\cite{kong2020physgan} introduced methods to generate physical-world-resilient adversarial examples that remain effective despite changing viewpoints and lighting, highlighting the practical urgency of robust autonomous systems. It is important to distinguish our study of abstract state-vector attacks from the extensive body of work targeting the upstream perception pipeline in the physical domain. A recent survey by Chi et al.~\cite{chi2024adversarial} provides a comprehensive taxonomy of these physical threats. Foundational work by Zhou et al.~\cite{zhou2020deepbillboard} demonstrated that physical billboards could be systematically manipulated to mislead steering decisions. Similarly, attacks have been developed for specific perception modules; for instance, Sato et al.~\cite{sato2021robustness} exposed the vulnerability of lane detection models to physical patches, directly affecting trajectory planning.

More recently, Rossolini et al.~\cite{rossolini2024real} extended the robustness evaluation of segmentation models against adversarial patches in realistic driving scenarios \cite{nesti2022evaluating} by analyzing the real-world robustness of real-time segmentation models, further confirming that sensor-level perception remains a critical attack surface. Unlike these works, which exploit the high-dimensional input space of sensors (e.g., cameras), our work assumes the perception layer has been bypassed or compromised and focuses on the vulnerability of the low-dimensional state vector itself.

In this paper, we construct a structured offline trajectory-learning framework based on real-world intersection driving data and systematically compare three learning paradigms: multilayer perceptron-based Behavior Cloning, Transformer-based Behavior Cloning with object tokenization \cite{vaswani2017attention}, and offline adversarial imitation learning in a GAIL-style formulation. We then evaluate these models under clean and adversarial perturbations such as Fast Gradient Sign
Method (FGSM) and PGD, across different crossing scenarios as shown in Fig. \ref{fig:Crossings}, to expose differences in robustness that are not captured by nominal trajectory accuracy metrics alone.

The contributions of this paper are:

\begin{itemize}
    \item An offline trajectory-learning pipeline based on structured real-world driving data.
    \item A controlled comparison of BC (MLP), BC (Transformer), and offline GAIL-style IRL under identical state and supervision contracts.
    \item A systematic inference-time adversarial robustness evaluation using FGSM and PGD across day and night conditions.
    \item An empirical analysis demonstrating that comparable nominal trajectory accuracy can conceal substantially different adversarial robustness profiles.
\end{itemize}

% %%%%%%%%%%%%%%%%%%%%%%%%%%%%%%%%%%%%%%%%%%%%%%%%%%%%%%%%%%%%%%%%%%%%%%%%
\section{Background and Related Work}

\subsection{Behavior Cloning and Imitation Learning}

Imitation learning formulates policy learning as supervised regression from expert demonstrations. In autonomous driving, Behavior Cloning (BC) maps observed states directly to expert actions or future trajectories \cite{codevilla2018conditional} and has been widely adopted due to its simplicity and scalability \cite{pomerleau1989alvinn,codevilla2019exploring,hussein2017imitation, bojarski2016end}.

Given a dataset $\mathcal{D}=\{(s_t, \tau_t^{\text{exp}})\}$ of state–trajectory pairs, BC learns a parametric predictor $\pi_\theta$ by minimizing:

\begin{equation}
\mathcal{L}_{\text{BC}}(\theta) =
\mathbb{E}_{(s_t, \tau_t^{\text{exp}})\sim \mathcal{D}}
\left[
\ell\big(\pi_\theta(s_t), \tau_t^{\text{exp}}\big)
\right],
\end{equation}

where $\ell(\cdot)$ denotes a regression loss (mean squared error or SmoothL1)
applied to the entire $H=20$ step future trajectory.

Despite strong nominal accuracy, BC is sensitive to distributional shift: small prediction errors can compound over time when policies operate outside the training distribution \cite{ross2011dagger, dehaan2019causal}. In safety-critical domains such as driving, such instability is especially concerning.

\subsection{Inverse Reinforcement Learning and GAIL}

Inverse Reinforcement Learning (IRL) seeks to recover a reward function that explains expert behavior rather than directly imitating actions \cite{ng2000algorithms}. Generative Adversarial Imitation Learning (GAIL) \cite{ho2016generative} reformulates IRL as a minimax game between a policy and a discriminator:

\begin{equation}
\min_{\pi_\theta} \max_{D_\phi}
\mathbb{E}_{\tau \sim \pi_E}[\log D_\phi(s,a)]
+
\mathbb{E}_{\tau \sim \pi_\theta}[\log(1 - D_\phi(s,a))].
\end{equation}

GAIL has demonstrated strong performance in continuous control domains by implicitly learning reward structures from demonstrations. However, adversarial objectives may alter stability characteristics and sensitivity to perturbations, particularly in offline settings.

Recent work analyzing robustness and stability in deep reinforcement learning highlights that learned policies can exhibit brittle behavior under structured perturbations \cite{mohan2026toward,mohan2025advancing,karpenahalli2025evolution}. Yet systematic comparisons between supervised BC and adversarial imitation under controlled trajectory-learning settings remain limited.

\subsection{Adversarial Attacks: FGSM and PGD}

Adversarial attacks construct worst-case perturbations that maximize prediction error \cite{szegedy2014intriguing, carlini2017towards}. The FGSM \cite{goodfellow2015explaining, moosavi2016deepfool} computes a single-step perturbation:

\begin{equation}
\tilde{s}_t = s_t + \epsilon \cdot \text{sign}(\nabla_{s_t} \mathcal{L}),
\end{equation}

while PGD \cite{madry2019towards} performs iterative gradient-based updates within an $\epsilon$-bounded region.

Although adversarial robustness has been extensively studied in image classification and reinforcement learning, including ensemble-based defenses and stability analyses \cite{mohan2025advancing,mohan2026toward}, its implications for structured offline trajectory predictors in autonomous driving remain underexplored.

This work bridges that gap by evaluating BC and offline GAIL-style trajectory models under controlled FGSM and PGD perturbations in real-world driving data. 

%%%%%%%%%%%%%%%%%%%%%%%%%%%%%%%%%%%%%%%%%%%%%%%%%%%%%%%%%%%%%%%%%%%%%%%%
\section{Experimental Setup}

\subsection{Vehicle Platform and Sensor Configuration}
\label{sec:veh}

Data were collected and evaluated using a full-scale Audi Q8 test vehicle equipped with a multi-sensor perception stack and an Autoware-based software framework \cite{autoware}. The platform includes a centrally mounted 3D LiDAR sensor, six cameras providing front, rear, and lateral coverage, and a high-precision GNSS/IMU (ADMA) unit for ego-state estimation.
All sensors are temporally synchronized within the vehicle coordinate frame as described in \cite{sekaran2025urbaning} and \cite{rossle2026driving}.
% All sensors are temporally synchronized within the vehicle coordinate frame shown in Fig.~2. 

While the trajectory-learning models are trained offline using recorded driving datasets, all inference-time robustness evaluations are performed by deploying the trained policies for \emph{open-loop real-time inference} on the test vehicle. During vehicle operation, structured state representations are constructed online from live Autoware perception \cite{autoware}, localization, and map-based annotation outputs. The learned policies generate trajectory predictions in real time; however, these predictions are \emph{not} fed back to the vehicle control stack and do not influence vehicle actuation.

This open-loop deployment enables realistic evaluation of inference-time robustness under real-world sensing, synchronization, and annotation conditions, while ensuring safe operation of the vehicle. An overview of the complete evaluation workflow is shown in Fig.~\ref{fig:Story}.

\begin{figure*}[t]
\centering
\includegraphics[width=\linewidth]{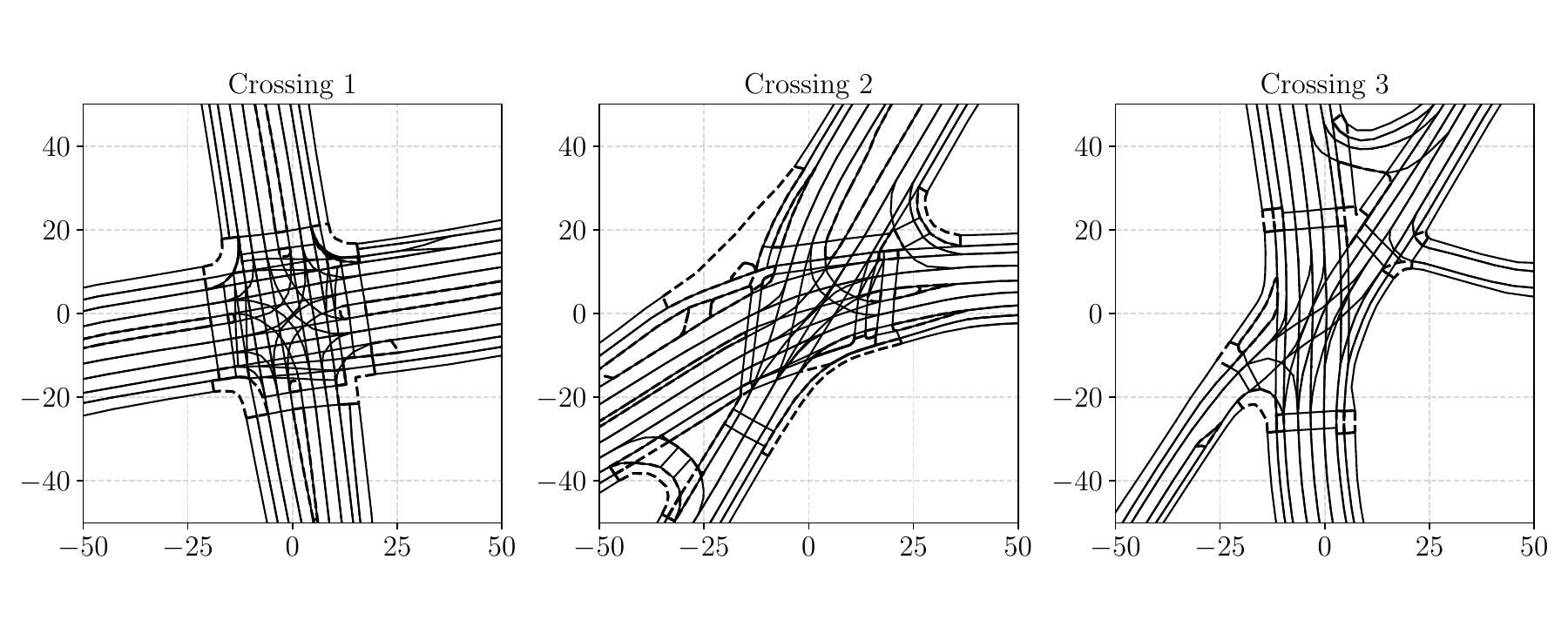}
\caption{All three crossings selected to evaluate and test the algorithms in real time.}
\label{fig:Crossings}
\end{figure*}

% \begin{figure}[t]
% \centering
% \includegraphics[width=0.8\linewidth]{map_vertical.pdf}
% \caption{All three crossings selected to evaluate and test the algorithms in real time.}
% \label{fig:Crossings}
% \end{figure}

\subsection{Dataset and Sequence Structure}
\label{sec:dataset}

The sensor data collected on the vehicle platform described in \ref{sec:veh} are organized and curated using the UrbanIng-V2X dataset~\cite{sekaran2025urbaning}, which provides temporally synchronized and spatially calibrated multi-sensor recordings across multiple urban intersections in Ingolstadt, Germany. The dataset includes vehicle-mounted sensing, high-definition maps, and consistent traffic participant annotations, enabling reproducible offline learning and evaluation.

Although UrbanIng-V2X supports multi-vehicle and infrastructure-based cooperative perception, this work intentionally focuses on a \emph{single-ego trajectory learning} setting to isolate policy robustness effects under controlled conditions. For each recorded sequence, one vehicle is designated as the ego agent, and ego motion signals, lane-level geometry, surrounding traffic participants, and map-based annotations are extracted from the synchronized logs and HD map representations.

The dataset is organized into intersection-level driving sequences, each spanning approximately 20 seconds at a nominal frequency of 10~Hz. From these sequences, temporally aligned samples are generated, forming state--trajectory pairs $(s_t, \tau_t)$ at each valid timestamp. Only samples satisfying strict future-horizon validity and annotation consistency are retained to ensure supervision integrity and reproducibility.

For completeness, we note that the same vehicle platform and annotation pipeline also support large-scale route-level data collection as described in the DrivIng dataset~\cite{rossle2026driving}. However, all experiments in this paper are conducted exclusively on intersection-centric sequences derived from UrbanIng-V2X.

These curated sequences provide the raw inputs from which the structured state representation is constructed, as described next in \ref{sec:State}.

\subsection{Structured State Construction}
\label{sec:State}

Raw driving logs are indexed and temporally aligned across all sensor streams. 
For each valid timestamp $t$, we construct a structured state vector 
$s_t \in \mathbb{R}^{97}$ by concatenating ego-level dynamics, lane-relative 
geometric features, and a fixed-size set of surrounding traffic agents. 
All quantities are expressed in the ego-centric coordinate frame unless stated otherwise.

\begin{itemize}
    \item \textbf{Ego dynamics (4D).} 
    The ego state encodes instantaneous vehicle motion as
    longitudinal speed $v$, longitudinal acceleration $a$, yaw rate $\dot{\psi}$, 
    and absolute heading $\psi$.

    \item \textbf{Lane-relative geometry (3D).} 
    Road alignment is represented using lane-centric quantities:
    lateral displacement from the lane centerline $e_y$, heading error relative 
    to the lane tangent $e_\psi$, and local lane curvature $\kappa$.

    \item \textbf{Surrounding objects (10 $\times$ 8D).} 
    The $K=10$ nearest traffic participants within a fixed radius of $60\,\mathrm{m}$ 
    are selected and ordered by increasing Euclidean distance to the ego vehicle. 
    Each object is encoded using an 8-dimensional feature vector:
    relative longitudinal position $x_{\mathrm{rel}}$, relative lateral position 
    $y_{\mathrm{rel}}$, relative yaw $\psi_{\mathrm{rel}}$, relative speed 
    $v_{\mathrm{rel}}$, object length, object width, semantic class identifier, 
    and Euclidean distance to the ego vehicle.  
    All relative quantities are defined in the ego coordinate frame. 
    If fewer than $K$ objects are present, remaining slots are zero-padded.

    \item \textbf{Object validity mask (10D).} 
    A binary mask indicates whether each object slot corresponds to a real 
    traffic participant or padded placeholder, enabling the model to distinguish 
    missing objects from valid measurements.
\end{itemize}
% The resulting state vector has dimensionality 
% $4 + 3 + (10 \times 8) + 10 = 97$.

The supervision target is a future trajectory
\[
\tau_t \in \mathbb{R}^{20 \times 4},
\]
representing 20-step predictions of $(\Delta x, \Delta y, \Delta \psi, v)$ in the local frame.

Only supervision-consistent samples satisfying strict future-horizon validity are retained during dataset construction to ensure reproducibility and alignment integrity.

\subsection{Model Architectures}

We evaluate three trajectory-learning paradigms under identical state and supervision contracts:

\paragraph{BC-MLP}
The behavior cloning multilayer perceptron (BC-MLP) is a feedforward neural
network that maps the 97-dimensional structured state vector to a future
trajectory representation. The network comprises two hidden layers with
512 units each and ReLU activations, followed by a linear output layer that
produces an 80-dimensional vector, which is reshaped into a $20 \times 4$
multi-step future ego trajectory.

\paragraph{BC-Transformer}
The Transformer-based behavior cloning model operates on a tokenized state
representation consisting of ego and lane tokens, $K{=}10$ surrounding-object
tokens, and an object-validity mask token. A \emph{classification (CLS) token}
is prepended to the token sequence to act as a learnable global aggregation
token, resulting in a total of 14 tokens. These tokens are processed by a
Transformer encoder with 4 layers, 8 attention heads, a model dimension of 192,
a feedforward dimension of 768, and a dropout rate of 0.1. A regression head
maps the encoded representation to an 80-dimensional output, which is reshaped
into a $20 \times 4$ predicted trajectory.

\paragraph{Offline IRL (GAIL-style)}
In the GAIL-style inverse reinforcement learning formulation, the policy network
shares the same architecture as the BC-MLP to ensure a controlled comparison
across learning paradigms. The discriminator comprises separate state and
trajectory branches, each with a hidden dimension of 256, whose outputs are
combined to produce a binary classification logit that distinguishes expert
from policy-generated trajectories.

All models share identical input representations and prediction horizons to enable controlled comparison.

\subsection{Training Protocol}

Algorithm~\ref{alg:training} summarizes the offline training procedure.

\begin{algorithm}[t]
\caption{Offline Trajectory Learning}
\label{alg:training}
\begin{algorithmic}[1]
\STATE Construct dataset $\mathcal{D} = \{(s_t, \tau_t)\}$
\FOR{each epoch}
    \FOR{each mini-batch $(s_t, \tau_t)$}
        \STATE Predict $\hat{\tau}_t = \pi_\theta(s_t)$
        \STATE Compute supervised loss $\mathcal{L}_{BC}$
        \IF{IRL model}
            \STATE Update discriminator $D_\phi$
            \STATE Compute adversarial generator loss
        \ENDIF
        \STATE Update policy parameters $\theta$
    \ENDFOR
\ENDFOR
\end{algorithmic}
\end{algorithm}

Optimization uses SmoothL1 loss over the full prediction horizon \cite{kingma2014adam}. IRL models incorporate an adversarial objective term and optional trajectory smoothness regularization.

% \subsection{Adversarial Robustness Evaluation}

% Inference-time robustness is evaluated by perturbing the structured state vector $s_t$ using gradient-based attacks (Section 2). Perturbations are applied under bounded $\ell_\infty$ constraints.

% We define a structured 3D evaluation matrix spanning model family, attack type, and illumination condition. The complete evaluation protocol is shown in Table~\ref{tab:ablation-3d}.

\subsection{Adversarial Robustness Evaluation}

Inference-time robustness is evaluated by perturbing the structured state vector $s_t$ using gradient-based attacks (Section~II). Perturbations are applied under bounded $\ell_\infty$ constraints.

All robustness evaluations are conducted by deploying the trained policies for \emph{open-loop real-time inference} on the test vehicle. During vehicle operation, structured states are constructed online from live ROS/Autoware topics, including ego dynamics, lane-relative geometry, and surrounding-object annotations. The predicted trajectories are evaluated against ground-truth future trajectories derived from logged vehicle data and annotations. Policy outputs are not fed back into the vehicle control stack.

We evaluate inference-time robustness using a structured three-dimensional protocol that jointly varies the model family (BC-MLP, BC-Transformer, IRL), the attack type (Clean, FGSM, PGD), and the intersection scenario (Crossing1, Crossing2, Crossing3) as shown in Fig. \ref{fig:Crossings}, yielding a total of 27 evaluation configurations. For each configuration, we report Average Displacement Error (ADE), and Final Displacement Error (FDE). Algorithm~\ref{alg:evaluation} outlines the robustness evaluation procedure.

\begin{algorithm}[t]
\caption{Inference-Time Robustness Evaluation}
\label{alg:evaluation}
\begin{algorithmic}[1]
\FOR{each trained model $m$}
    \FOR{each attack $a \in \{\text{Clean, FGSM, PGD}\}$}
        \FOR{each crossing $c \in \{\text{1, 2, 3}\}$}
            \STATE Acquire synchronized live messages from ROS/Autoware topics during vehicle operation
            \STATE Build structured state $s_t$ and target $\tau_t$ using map + annotations
            \IF{$a \neq$ Clean}
                \STATE Apply perturbation to $s_t$ (FGSM/PGD)
            \ENDIF
            \STATE Predict $\hat{\tau}_t = \pi_m(s_t)$
            \STATE Compute ADE and FDE between $\hat{\tau}_t$ and $\tau_t$
        \ENDFOR
    \ENDFOR
\ENDFOR
\end{algorithmic}
\end{algorithm}

\subsection{Adversarial Attack Protocol}
Inference-time adversarial robustness is evaluated using FGSM and PGD attacks applied directly to the structured state vector $s_t \in \mathbb{R}^{97}$. We use an $\ell_\infty$-bounded perturbation budget $\epsilon = 0.05$. For PGD, we perform 10 iterative steps with step size $\alpha = 0.01$ and project intermediate states back to the $\ell_\infty$ ball $[s_t-\epsilon, s_t+\epsilon]$. Attacks are applied to all state dimensions, including ego, lane, object features, and mask entries. Gradients are computed with respect to a mean squared error loss
on the predicted XY trajectory over the full prediction horizon. Because the state contains heterogeneous physical quantities, attacks operate in the raw numeric feature space without per-feature normalization.

%%%%%%%%%%%%%%%%%%%%%%%%%%%%%%%%%%%%%%%%%%%%%%%%%%%%%%%%%%%%%%%%%%%%%%%%
\section{Results and Discussion}

This section analyzes inference-time adversarial robustness of offline trajectory-learning models trained on real-world intersection data. Quantitative results are reported using trajectory-level metrics, while qualitative analysis is used to characterize dominant failure modes under adversarial perturbations. All comparisons are conducted under identical state representations, supervision horizons, and evaluation protocols. Both ADE and FDE are reported to distinguish average trajectory deviation from terminal error accumulation, which is particularly pronounced under iterative adversarial attacks.

\subsection{Nominal Trajectory Prediction Accuracy}

Under clean (non-adversarial) conditions, all evaluated models achieve low trajectory prediction error across intersections. As reported in Table~\ref{tab:ablation_grouped}, ADE and FDE values remain below $0.08$ for BC-MLP, BC-Transformer, and IRL models in all crossings.

These results indicate that, given the structured state representation and fixed prediction horizon, each model family is capable of accurately fitting the offline trajectory-learning objective. Importantly, the similarity in clean-condition performance establishes that differences observed under adversarial perturbations are not attributable to underfitting, data scarcity, or architectural capacity, but rather to differences in sensitivity of the learned input--output mappings.

%---------------------------------------Plots-------------------------------------------------

\begin{figure*}[t]
\centering

% ---------- Row 1 ----------
\subfloat[\scriptsize BC-T (FGSM)]{\includegraphics[width=0.323\textwidth]{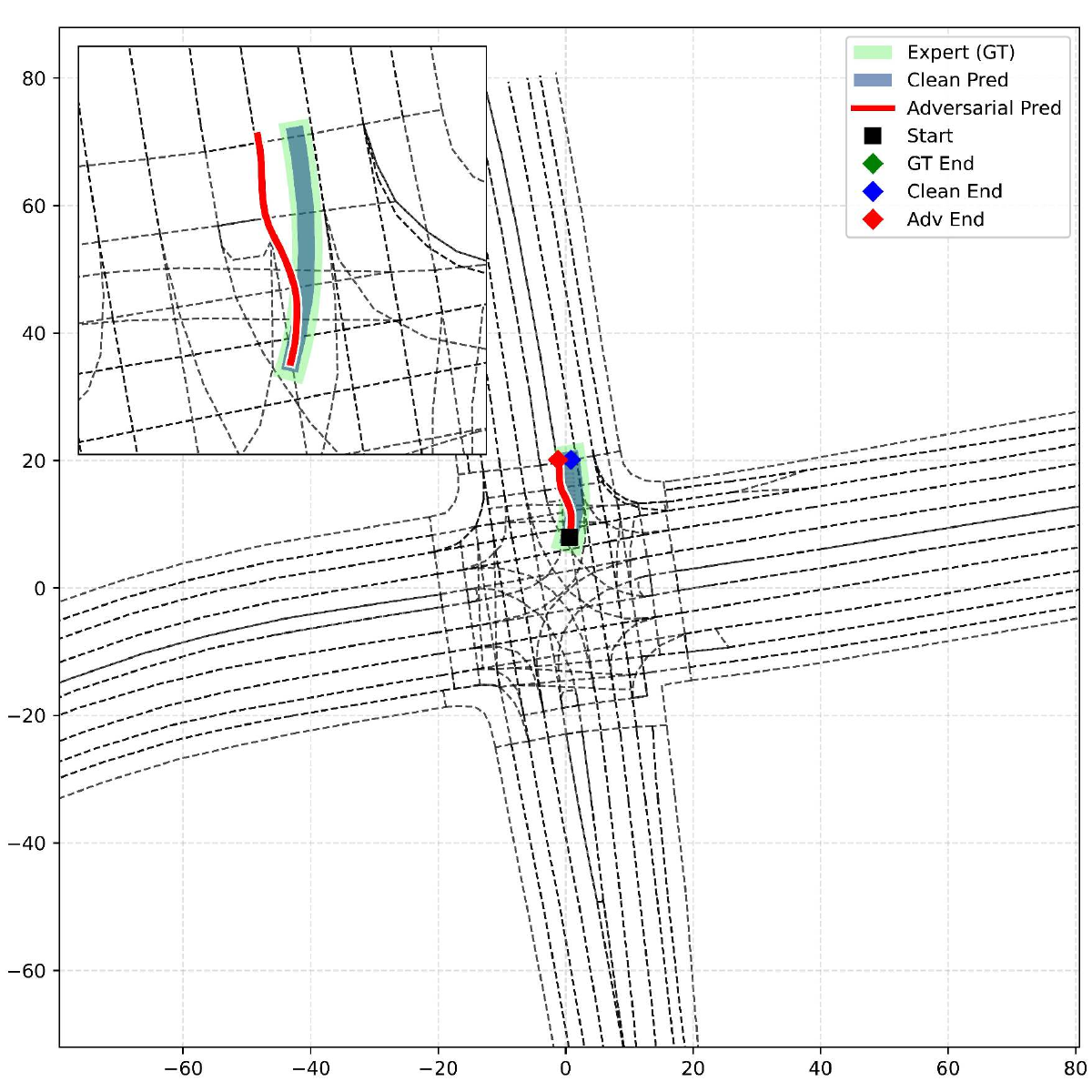}}
\hspace{0.8mm}
\subfloat[\scriptsize IRL (FGSM)]{\includegraphics[width=0.323\textwidth]{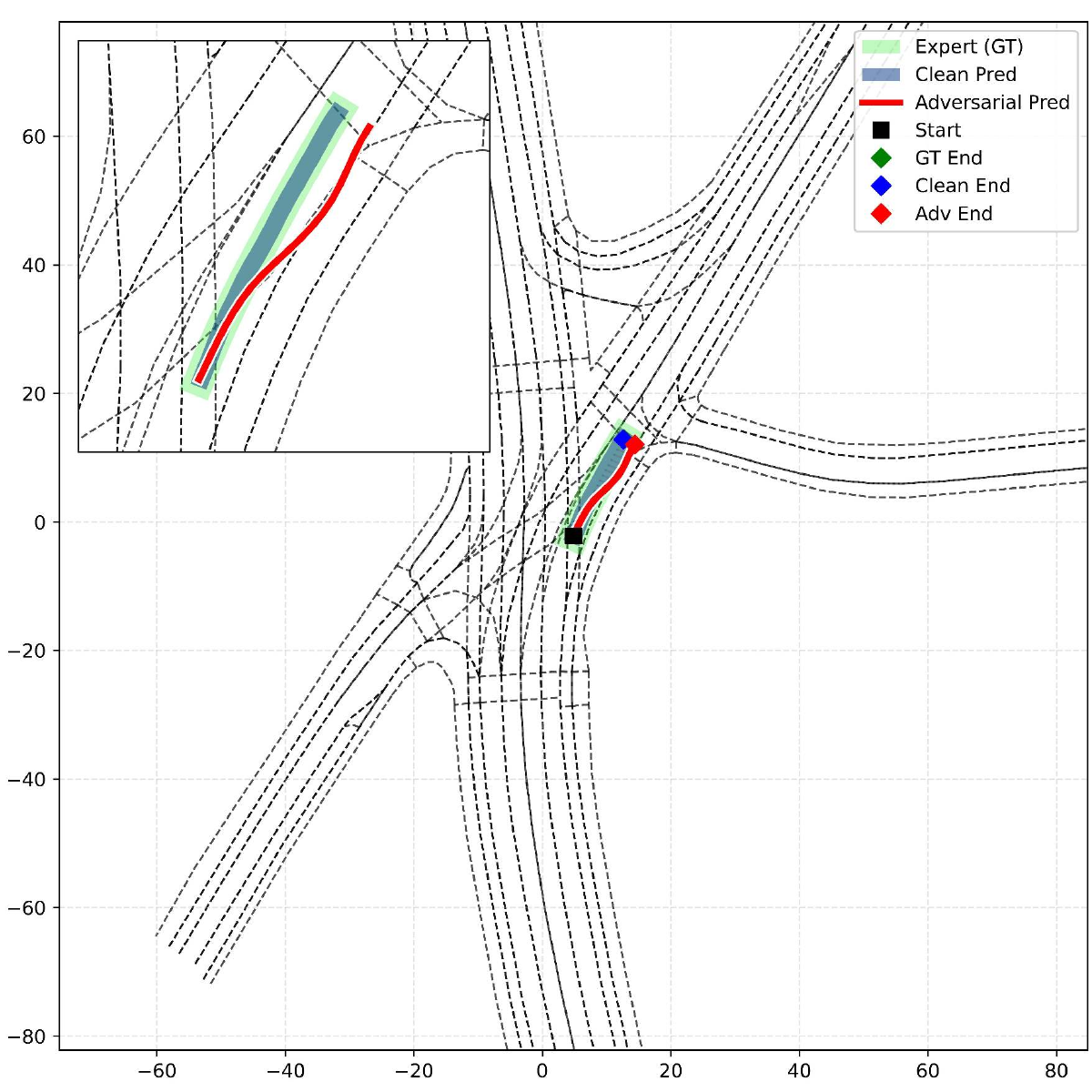}}
\hspace{0.8mm}
\subfloat[\scriptsize BC-MLP (FGSM)]{\includegraphics[width=0.323\textwidth]{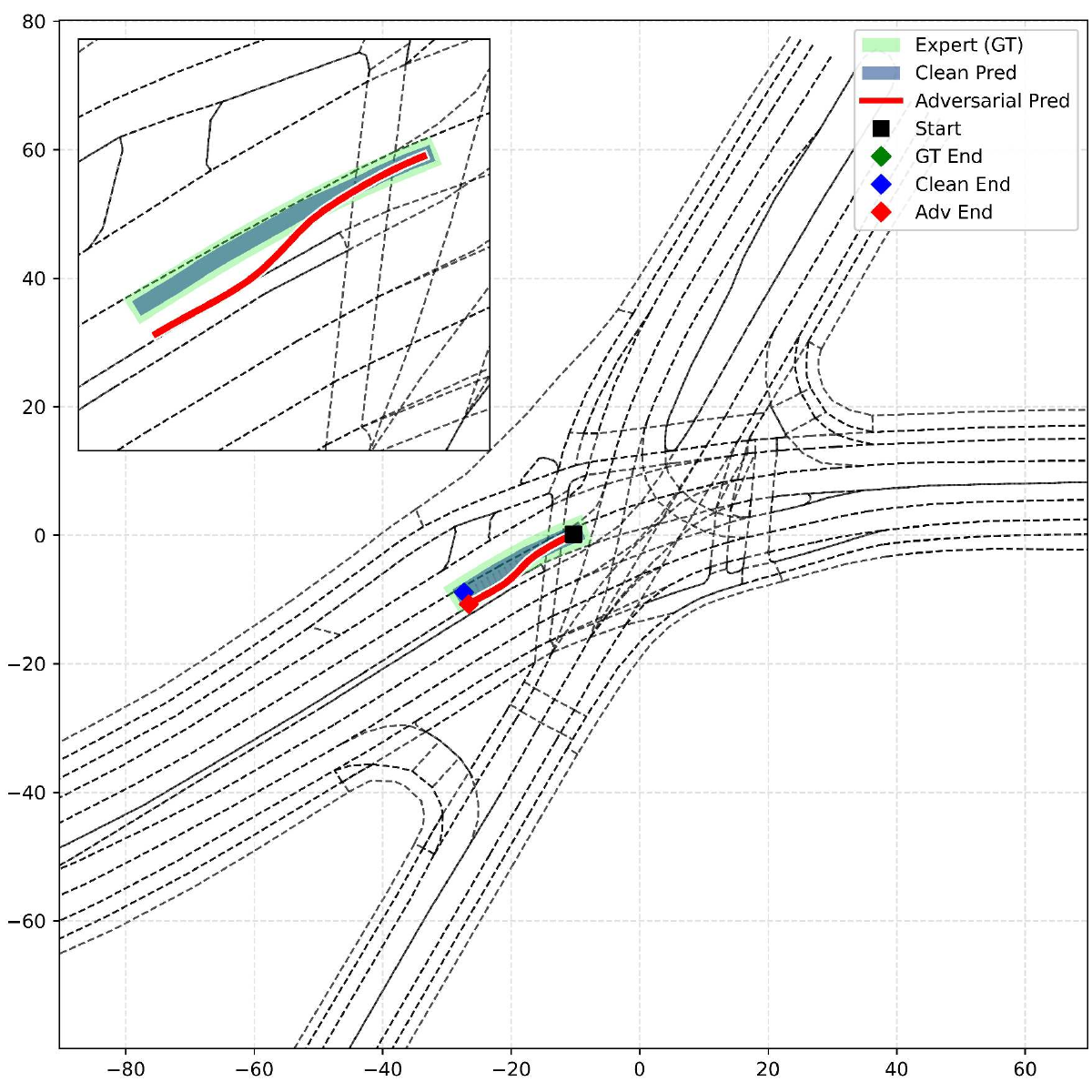}}

\vspace{0.6mm}

% ---------- Row 2 ----------
\subfloat[\scriptsize BC-MLP (FGSM)]{\includegraphics[width=0.323\textwidth]{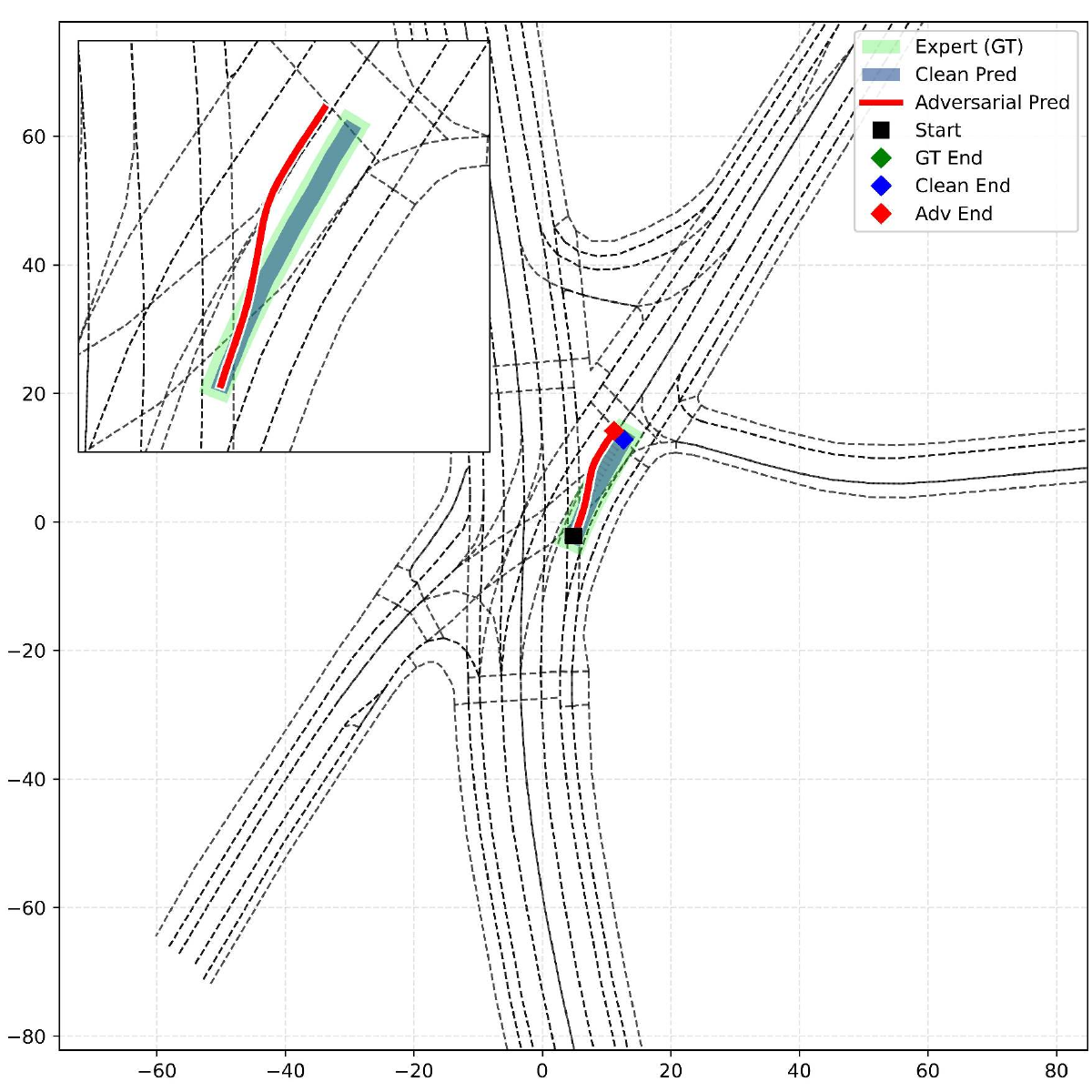}}
\hspace{0.8mm}
\subfloat[\scriptsize BC-T (PGD)]{\includegraphics[width=0.323\textwidth]{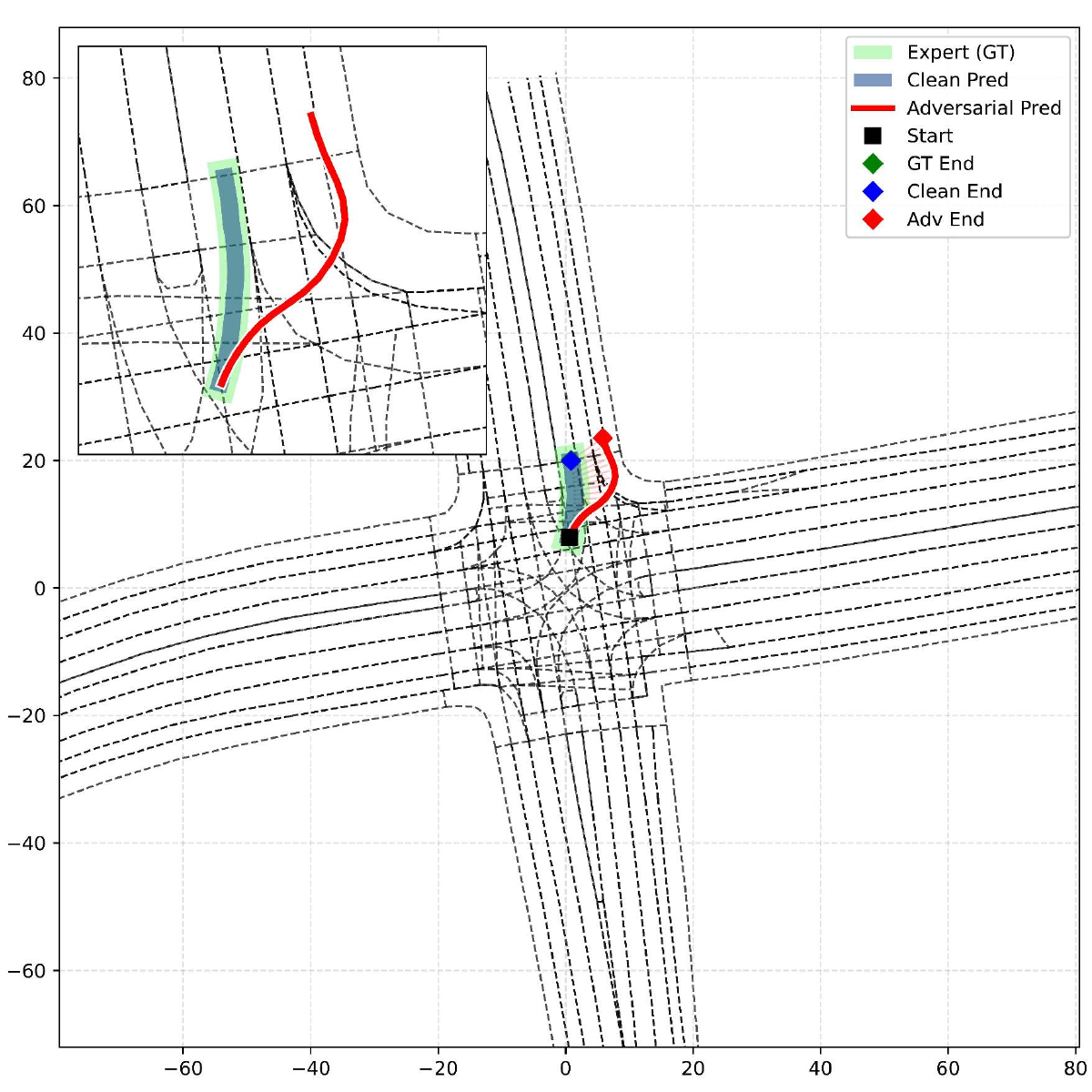}}
\hspace{0.8mm}
\subfloat[\scriptsize BC-MLP (PGD)]{\includegraphics[width=0.323\textwidth]{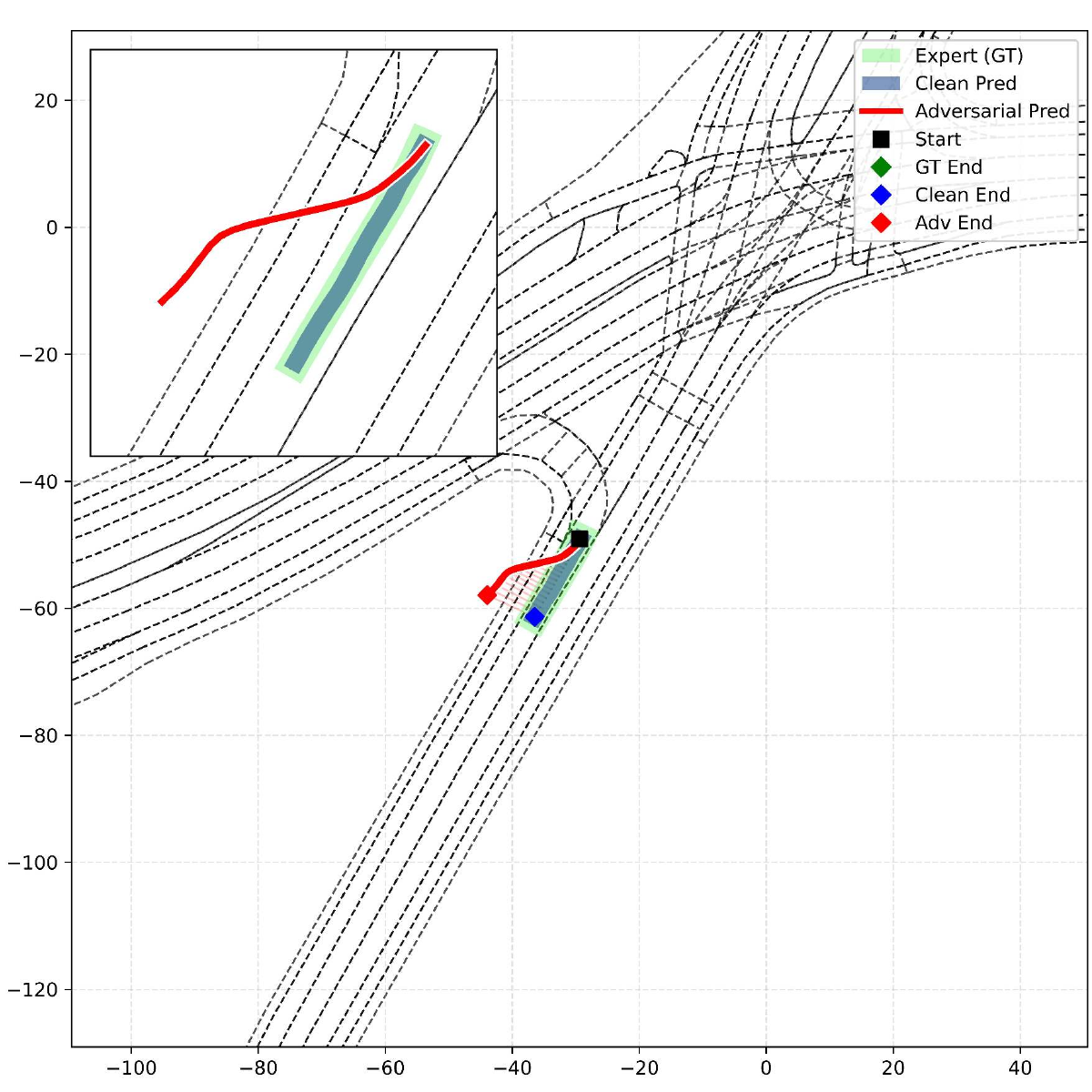}}

\vspace{0.6mm}

% ---------- Row 3 ----------
\subfloat[\scriptsize BC-MLP (PGD)]{\includegraphics[width=0.323\textwidth]{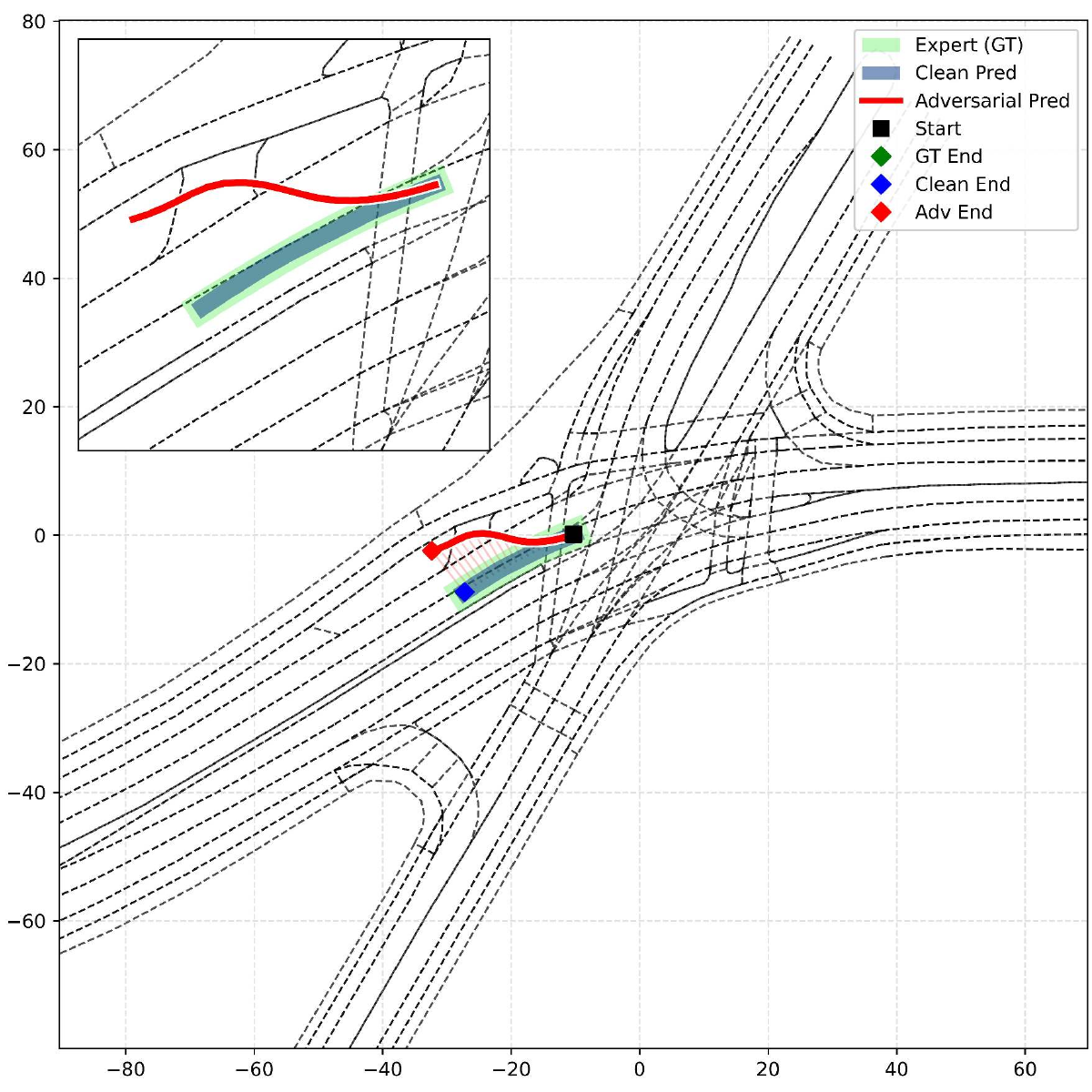}}
\hspace{0.8mm}
\subfloat[\scriptsize IRL (PGD)]{\includegraphics[width=0.323\textwidth]{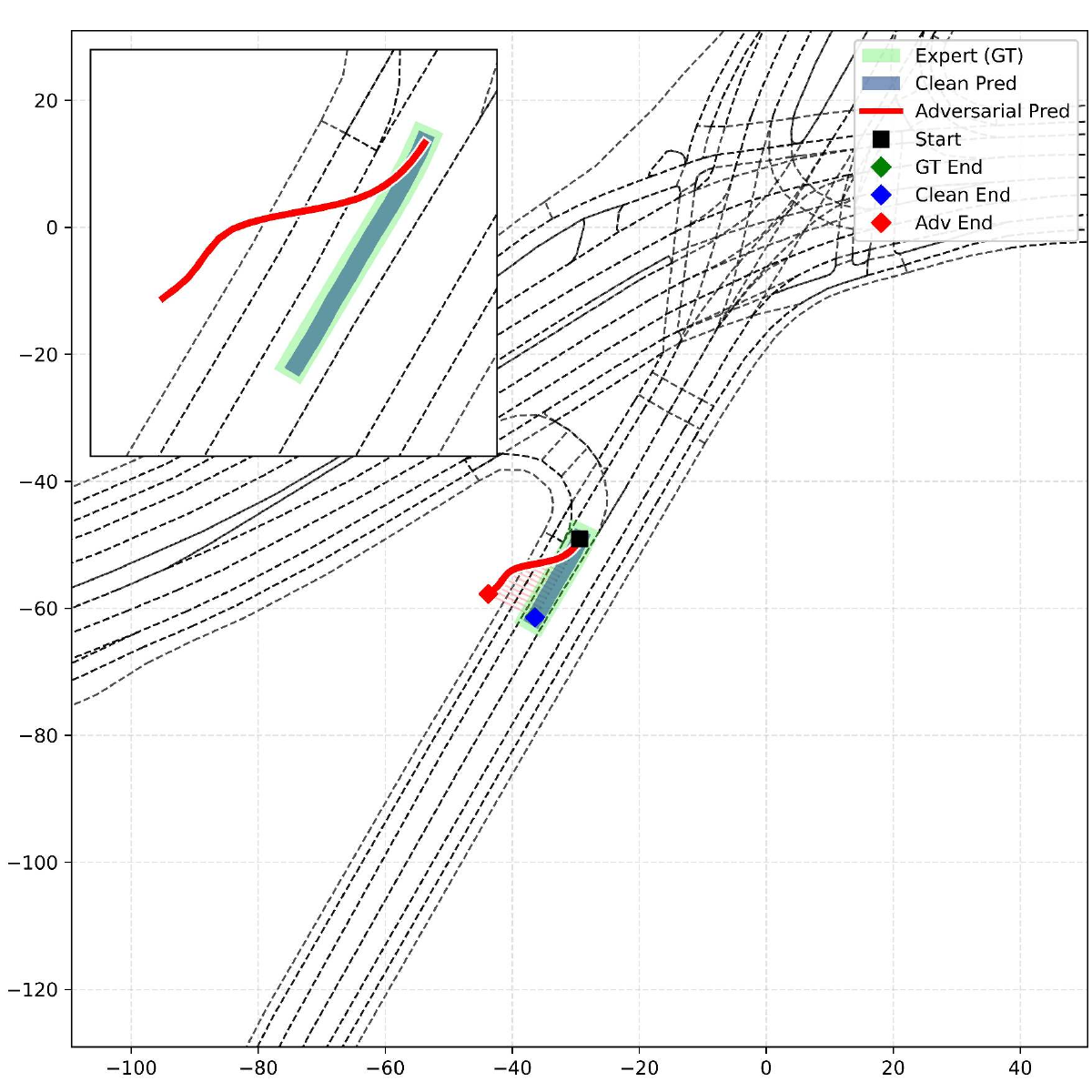}}
\hspace{0.8mm}
\subfloat[\scriptsize BC-T (PGD)]{\includegraphics[width=0.323\textwidth]{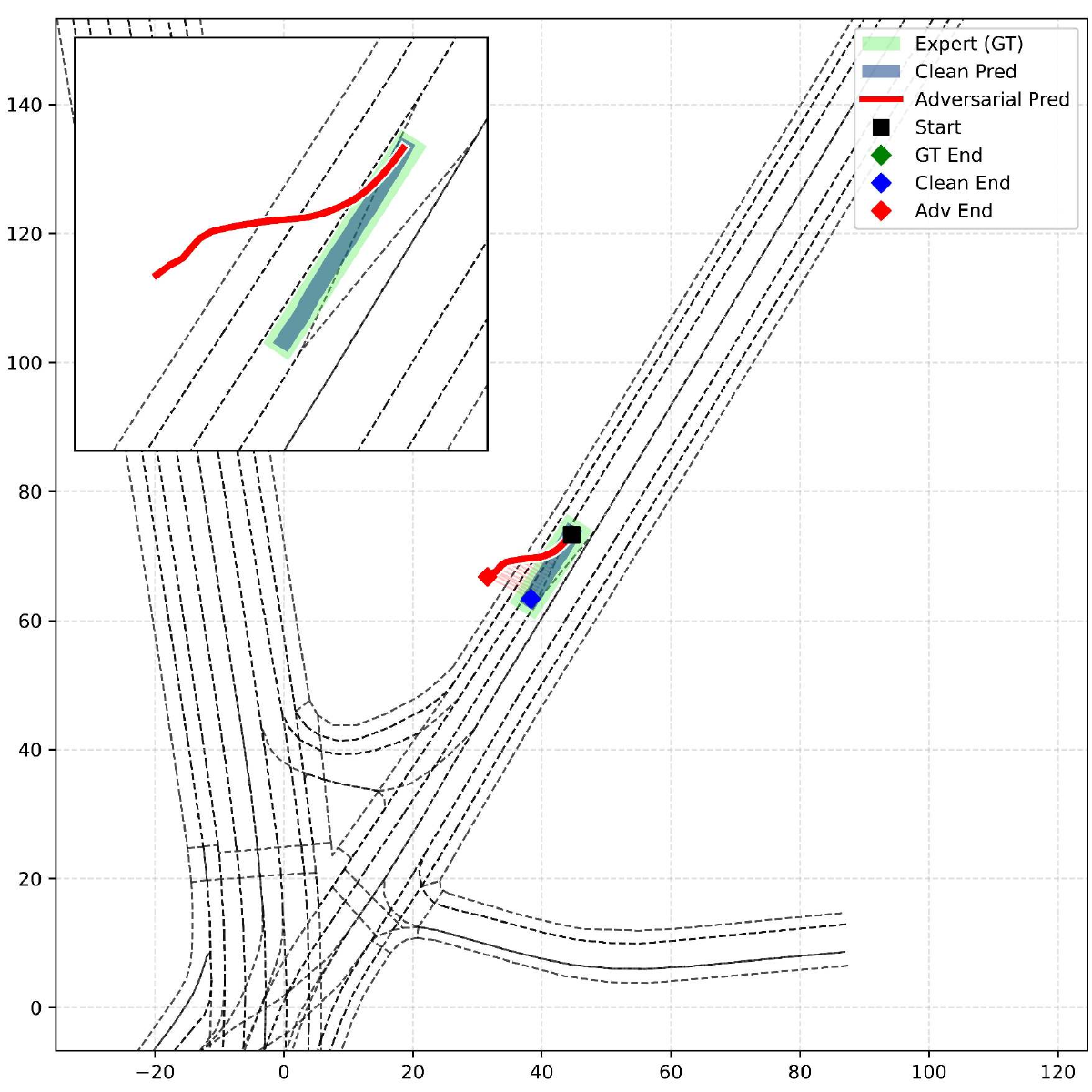}}

\caption{\textbf{Top-9 severe adversarial failures across real-world driving scenarios.}
The four most severe FGSM cases (top two rows, left-to-right) and the five most severe PGD cases (bottom row) are shown, ranked by mean $\Delta$FDE.
Each tile visualizes the expert trajectory (green), the clean policy prediction (blue), and the adversarially perturbed prediction (red).
The corresponding policy architecture (BC-MLP, BC-Transformer, IRL) and the attack type are indicated below each subplot.}
\label{fig:top9_severe_tiles}

\end{figure*}

% ====================Table (updated)==============
\begin{table*}[]
\centering
\caption{\textbf{Grouped inference-time adversarial ablation across all crossings.} \textbf{Bold} values indicate the best (lowest) robustness metrics within each crossing and attack setting}
\label{tab:ablation_grouped}
\begin{tabular}{lllrrrrrr}
\hline
Crossing & Model & Attack & ADE$_{\text{clean}}\!\downarrow$ & FDE$_{\text{clean}}\!\downarrow$ & ADE$_{\text{adv}}\!\downarrow$ & FDE$_{\text{adv}}\!\downarrow$ & $\Delta$ADE$\downarrow$ & $\Delta$FDE$\downarrow$ \\ \hline
\multicolumn{1}{c}{\multirow{9}{*}{crossing1}} & \multirow{3}{*}{bc\_mlp} & clean & 0.049 & 0.048 & 0.049 & 0.048 & 0.000 & 0.000 \\
\multicolumn{1}{c}{} &  & fgsm & 0.056 & 0.056 & 1.125 & 1.898 & 1.070 & 1.842 \\
\multicolumn{1}{c}{} &  & pgd & 0.053 & 0.066 & \textbf{3.237} & \textbf{5.301} & \textbf{3.185} & \textbf{5.236} \\ \cline{2-9} 
\multicolumn{1}{c}{} & \multirow{3}{*}{bc\_transformer} & clean & 0.053 & 0.053 & 0.053 & 0.053 & 0.000 & 0.000 \\
\multicolumn{1}{c}{} &  & fgsm & 0.053 & 0.043 & \textbf{1.104} & \textbf{1.887} & \textbf{1.052} & 1.843 \\
\multicolumn{1}{c}{} &  & pgd & 0.066 & 0.077 & 3.713 & 6.804 & 3.647 & 6.727 \\ \cline{2-9} 
\multicolumn{1}{c}{} & \multirow{3}{*}{irl} & clean & 0.064 & 0.053 & 0.064 & 0.053 & 0.000 & 0.000 \\
\multicolumn{1}{c}{} &  & fgsm & 0.064 & 0.081 & 1.123 & 1.916 & 1.059 & \textbf{1.835} \\
\multicolumn{1}{c}{} &  & pgd & 0.058 & 0.054 & 3.603 & 5.845 & 3.545 & 5.792 \\ \hline
\multirow{9}{*}{crossing2} & \multirow{3}{*}{bc\_mlp} & clean & 0.058 & 0.056 & 0.058 & 0.056 & 0.000 & 0.000 \\
 &  & fgsm & 0.068 & 0.050 & 1.154 & 1.921 & 1.086 & 1.871 \\
 &  & pgd & 0.061 & 0.054 & 3.903 & 7.987 & 3.841 & 7.933 \\ \cline{2-9} 
 & \multirow{3}{*}{bc\_transformer} & clean & 0.058 & 0.051 & 0.058 & 0.051 & 0.000 & 0.000 \\
 &  & fgsm & 0.062 & 0.069 & \textbf{1.142} & \textbf{1.929} & \textbf{1.079} & \textbf{1.859} \\
 &  & pgd & 0.051 & 0.054 & \textbf{3.278} & \textbf{5.358} & \textbf{3.227} & \textbf{5.303} \\ \cline{2-9} 
 & \multirow{3}{*}{irl} & clean & 0.050 & 0.047 & 0.050 & 0.047 & 0.000 & 0.000 \\
 &  & fgsm & 0.057 & 0.051 & 1.516 & 2.951 & 1.459 & 2.900 \\
 &  & pgd & 0.060 & 0.060 & 3.899 & 7.966 & 3.839 & 7.905 \\ \hline
\multirow{9}{*}{crossing3} & \multirow{3}{*}{bc\_mlp} & clean & 0.052 & 0.059 & 0.052 & 0.059 & 0.000 & 0.000 \\
 &  & fgsm & 0.053 & 0.060 & 1.114 & \textbf{1.897} & 1.062 & \textbf{1.838} \\
 &  & pgd & 0.059 & 0.085 & 3.611 & 5.881 & 3.552 & 5.796 \\ \cline{2-9} 
 & \multirow{3}{*}{bc\_transformer} & clean & 0.064 & 0.078 & 0.064 & 0.078 & 0.000 & 0.000 \\
 &  & fgsm & 0.057 & 0.071 & \textbf{1.114} & 1.910 & \textbf{1.057} & 1.840 \\
 &  & pgd & 0.060 & 0.062 & 3.729 & 6.812 & 3.669 & 6.750 \\ \cline{2-9} 
 & \multirow{3}{*}{irl} & clean & 0.060 & 0.064 & 0.060 & 0.064 & 0.000 & 0.000 \\
 &  & fgsm & 0.054 & 0.060 & 1.126 & 1.920 & 1.072 & 1.860 \\
 &  & pgd & 0.066 & 0.059 & \textbf{3.241} & \textbf{5.273} & \textbf{3.175} & \textbf{5.214} \\ \hline
\end{tabular}
\end{table*}

% \subsection{Adversarial Degradation Patterns}

\subsection{Inference-Time Robustness Under FGSM Perturbations}

FGSM perturbations introduce significant degradation in trajectory accuracy across all models and crossings. Table~\ref{tab:ablation_grouped} shows that FGSM increases ADE by approximately $1.0$--$1.5$ meters and FDE by $1.8$--$2.9$ meters relative to clean predictions.

BC-MLP and BC-Transformer models exhibit comparable degradation magnitudes, indicating that increased architectural expressiveness and attention-based object tokenization do not substantially reduce vulnerability to single-step gradient-based perturbations. This suggests that FGSM exploits local linearity in the learned state-to-trajectory mapping that is shared across both architectures.
IRL-based models demonstrate slightly reduced ADE degradation in select crossings; however, this behavior is inconsistent and does not persist across all scenarios. The results suggest that reward-structured training may locally smooth prediction sensitivity in some state regions, but does not provide systematic robustness against adversarial perturbations applied at inference time.

\subsection{Robustness Degradation Under PGD Attacks}

PGD attacks result in substantially larger degradation than FGSM across all model families. As shown in Table~\ref{tab:ablation_grouped}, $\Delta$FDE frequently exceeds $5.0$ meters and reaches nearly $8.0$ meters in complex intersections, indicating severe deviation from expert trajectories.

BC-MLP exhibits the largest degradation under PGD, particularly in crossings with high curvature or merging traffic flows. This behavior reflects the sensitivity of fully connected architectures to iterative adversarial amplification, where small perturbations compound across successive gradient steps.

BC-Transformer models show modest reductions in degradation relative to BC-MLP in some configurations, but remain highly vulnerable overall. This indicates that attention mechanisms and object-level tokenization do not inherently constrain adversarial error accumulation when perturbations are repeatedly optimized.

IRL-based policies do not exhibit consistent robustness advantages under PGD. In several crossings, IRL degradation is comparable to or greater than that of BC-based models. This suggests that adversarial training objectives used during offline imitation do not translate into inference-time robustness when the policy is evaluated under state perturbations outside the training distribution.

\subsection{Scenario-Dependent Sensitivity}

Robustness degradation varies systematically across intersection scenarios. Crossings characterized by tighter curvature, lane merges, or complex traffic geometry consistently produce higher $\Delta$ADE and $\Delta$FDE values across all models.

These results indicate that adversarial vulnerability is jointly determined by model architecture and local state geometry. Small perturbations to lane-relative features or surrounding-object encodings can induce large downstream trajectory deviations in geometrically constrained scenarios, particularly near turning points and conflict zones.

This observation highlights the limitations of aggregate robustness metrics and motivates scenario-aware evaluation for safety-critical autonomous driving systems.

\subsection{Qualitative Failure Characteristics}

Fig.~\ref{fig:top9_severe_tiles} visualizes the most severe adversarial failures ranked by mean $\Delta$FDE. Under clean conditions, predicted trajectories closely follow expert demonstrations and respect lane geometry. Under adversarial perturbations, predicted trajectories frequently diverge from the intended path, exhibiting excessive curvature, lane departure, or overshooting of turn exits.

A notable characteristic of these failures is that adversarial trajectories often remain kinematically smooth while being semantically incorrect. Such trajectories may satisfy low-level motion constraints while violating high-level driving semantics, such as lane adherence or intersection rules.

In PGD cases, trajectory deviations typically increase over the prediction horizon, resulting in large terminal displacement errors. This explains the dominance of $\Delta$FDE over $\Delta$ADE in the quantitative results and underscores the importance of final-state metrics in robustness evaluation.

\subsection{Discussion}

The presented results demonstrate that nominal trajectory prediction accuracy is not indicative of adversarial robustness. Models with nearly identical clean performance exhibit substantially different degradation profiles under inference-time perturbations. Architectural inductive biases and training paradigms influence not only predictive accuracy but also the local geometry of the learned state-to-trajectory mapping, which directly affects adversarial sensitivity.

None of the evaluated approaches exhibits inherent robustness to gradient-based attacks, indicating that robustness must be explicitly addressed rather than assumed as a byproduct of model complexity or adversarial imitation objectives.

%%%%%%%%%%%%%%%%%%%%%%%%%%%%%%%%%%%%%%%%%%%%%%%%%%%%%%%%%%%%%%%%%%%%%%%%

\section{Conclusion}

This work presented a controlled offline evaluation framework for analyzing inference-time adversarial robustness of trajectory-learning policies in real-world autonomous driving scenarios. By leveraging structured state representations derived from real intersection driving data, we enabled a systematic comparison of behavior cloning with multilayer perceptrons, Transformer-based behavior cloning, and offline GAIL-style inverse reinforcement learning under identical state and supervision contracts.

Quantitative and qualitative results demonstrate that, despite comparable nominal trajectory prediction accuracy under clean conditions, all evaluated models exhibit substantial vulnerability to gradient-based adversarial perturbations. Both FGSM and PGD attacks induce significant trajectory deviations, with PGD causing severe final-state errors, particularly in geometrically complex intersections. These findings indicate that nominal performance metrics alone are insufficient for assessing the safety and robustness of offline trajectory-learning policies.

The analysis further reveals that architectural expressiveness and adversarial imitation objectives do not inherently confer robustness at inference time. Qualitative inspection of severe failure cases shows that adversarially perturbed trajectories can remain kinematically smooth while violating semantic driving constraints, highlighting a critical gap between clean-condition accuracy and safety-relevant behavior.

The proposed framework is modular, reproducible, and compatible with real-world autonomous driving datasets, providing a principled benchmark for studying adversarial robustness in offline trajectory-learning systems. It establishes a foundation for future work on robustness-aware model design and evaluation in safety-critical autonomous driving applications. In future work, we plan to investigate:
\begin{itemize}
    \item Robustness-oriented training strategies, including adversarially informed regularization and state-space smoothing.
    \item Scenario-aware robustness evaluation across a broader range of intersection geometries and traffic densities.
    \item Integration of robustness metrics with safety monitors and confidence-aware fallback mechanisms.
    \item Defense-aware training strategies and their robustness under closed-loop real-vehicle deployment.
    \item Feature-specific adversarial analysis, including attacks restricted
    to individual state components (e.g., ego dynamics, lane geometry, or
    surrounding-object features), to disentangle the relative vulnerability
    of different state subspaces.
    \item System-level latency and real-time feasibility analysis, including end-to-end inference latency, adversarial attack–generation overhead, and their impact on deployment in safety-critical autonomous driving systems, as well as the role of deployment-oriented optimizations (e.g., TensorRT-based inference acceleration).

\end{itemize}

Collectively, these directions aim to advance the development of robust, reliable, and deployable trajectory-learning systems that can withstand adversarial perturbations while meeting the stringent safety and real-time constraints of autonomous driving.

%%%%%%%%%%%%%%%%%%%%%%%%%%%%%%%%%%%%%%%%%%%%%%%%%%%%%%%%%%%%%%%%%%%%%%%%

%%%%%%%%%%%%%%%%%%%%%%%%%%%%%%%%%%%%%%%%%%%%%%%%%%%%%%%%%%%%%%%%%%%%%%%%

% \section{Citations and references}

%%%%%%%%%%%%%%%%%%%%%%%%%%%%%%%%%%%%%%%%%%%%%%%%%%%%%%%%%%%%%%%%%%%%%%%%

%%% Use this environment to include acknowledgements (optional).
%%% This will be omitted in doubleblind mode.

% \begin{ack}
% By using the \texttt{ack} environment to insert your (optional) 
% acknowledgements, you can ensure that the text is suppressed whenever 
% you use the \texttt{doubleblind} option. In the final version, 
% acknowledgements may be included on the extra page intended for references.
% \end{ack}

%%%%%%%%%%%%%%%%%%%%%%%%%%%%%%%%%%%%%%%%%%%%%%%%%%%%%%%%%%%%%%%%%%%%%%%%

%%% Use this command to include your bibliography file.

\bibliographystyle{IEEEtran}
\bibliography{mybibfile}
%%%%%%%%%%%%%%%%%%%%%%%%%%%%%%%%%%%%%%%%%%%%%%%%%%%%%%%%%%%%%%%%%%%%%%

\end{document}